\documentclass{egpubl}
\usepackage{pg2025s}

\usepackage{placeins}
\usepackage{amsmath}
\usepackage{booktabs}
\usepackage{colortbl}
\usepackage{graphicx}
\usepackage{xspace}
\usepackage{multirow}
\usepackage{makecell}
\usepackage{algorithm}
\usepackage{float}
\usepackage[normalem]{ulem}
\newcommand{\best}{\cellcolor{tablered}}
\newcommand{\sbest}{\cellcolor{orange}}
\newcommand{\tbest}{\cellcolor{yellow}}

\newcommand{\bc}{\mathbf{c}}

\newcommand{\bJ}{\mathbf{J}}

\newcommand{\bn}{\mathbf{n}}\newcommand{\bN}{\mathbf{N}}

\newcommand{\bp}{\mathbf{p}}

\newcommand{\bR}{\mathbf{R}}
\newcommand{\bS}{\mathbf{S}}
\newcommand{\bt}{\mathbf{t}}
\newcommand{\bu}{\mathbf{u}}

\newcommand{\bW}{\mathbf{W}}
\newcommand{\bx}{\mathbf{x}}

\newcommand{\bSigma}{\boldsymbol{\Sigma}}

\newcommand{\cG}{\mathcal{G}}

\newcommand{\cL}{\mathcal{L}}

\makeatletter
\DeclareRobustCommand\onedot{\futurelet\@let@token\@onedot}
\def\@onedot{\ifx\@let@token.\else.\null\fi\xspace}
\def\eg{e.g\onedot} 
\def\ie{i.e\onedot}

\makeatother

\definecolor{yellow}{rgb}{1, 1, 0.7}
\definecolor{orange}{rgb}{1, 0.85, 0.7}
\definecolor{tablered}{rgb}{1, 0.7, 0.7}
\definecolor{red}{rgb}{1, 0, 0}

\definecolor{wincolor}{rgb}{0.85, 0.0, 0.0}

\definecolor{darkyellow}{rgb}{0.8, 0.8, 0.5}
\definecolor{darkred}{rgb}{0.7, 0.3, 0.3}
\definecolor{darkgreen}{rgb}{0.3, 0.7, 0.3}
\definecolor{green}{rgb}{0, 1.0, 0}
\definecolor{pink}{rgb}{1, 0.4, 0.7}

\newcommand{\yang}[1]{#1}

\newcommand{\yyx}[1]{#1} 

\SpecialIssuePaper         

\CGFStandardLicense

\usepackage[T1]{fontenc}
\usepackage{dfadobe}  

\usepackage{cite}  
\BibtexOrBiblatex
\electronicVersion
\PrintedOrElectronic

\usepackage{egweblnk} 

\title[Unbiased 2DGS]%
      {Introducing Unbiased Depth into 2D Gaussian Splatting for High-accuracy Surface Reconstruction}

\author[Y. Yang et al.]{
\parbox{\textwidth}{\centering Yixin Yang \qquad Yang Zhou\thanks{Corresponding author.}  \qquad Hui Huang \\
\parbox{\textwidth}{\centering Visual Computing Research Center, CSSE, Shenzhen University }
}
}

\begin{document}

\teaser{
 \includegraphics[width=0.9\linewidth]{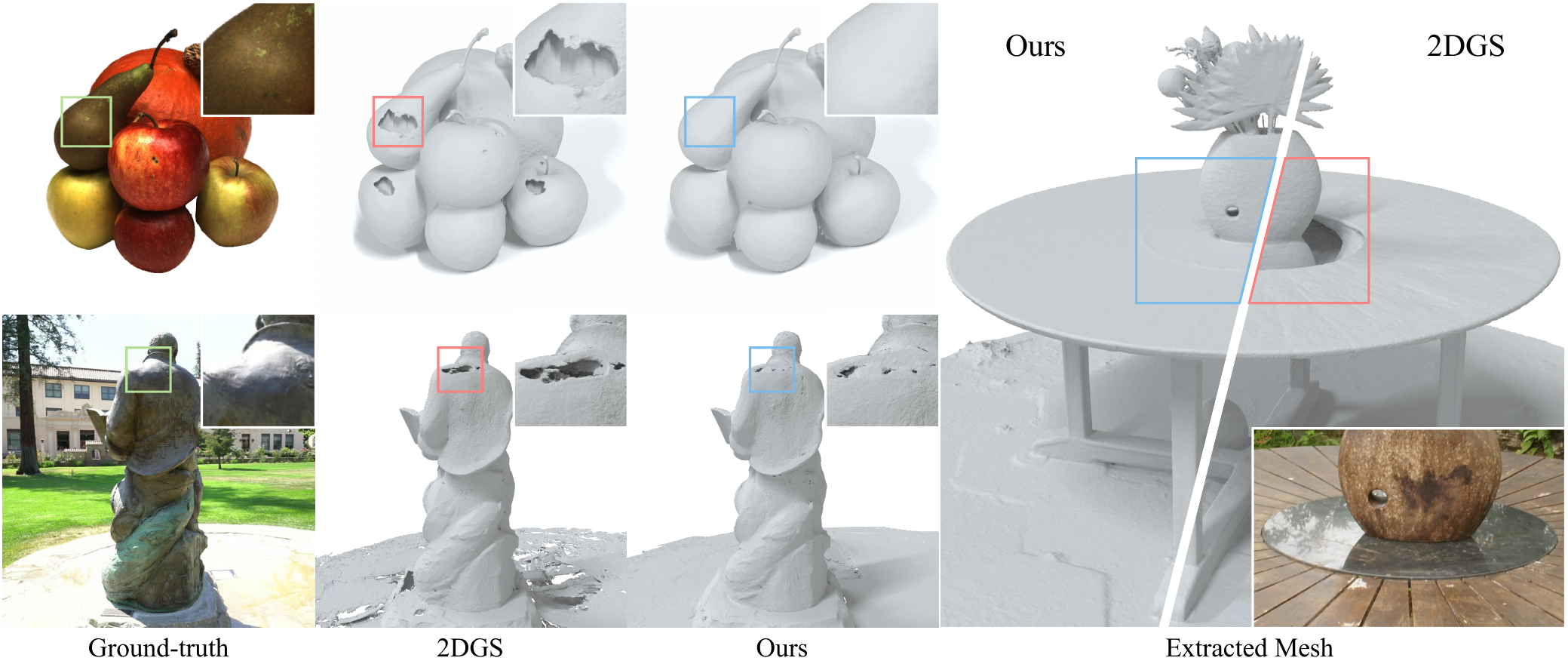}
 \centering
  \caption{As a severe limitation, 2D Gaussian splatting~\cite{huang20242d} has great difficulty handling high-specular regions, resulting in notable holes or pits in the reconstruction. Addressing this issue, we introduce unbiased depth into 2DGS, effectively solving the holes in specular highlight regions and significantly improving the geometry quality of surface reconstruction.}
\label{fig:teaser}
}

\maketitle
\begin{abstract}

Recently, 2D Gaussian Splatting (2DGS) has demonstrated superior geometry reconstruction quality than the popular 3DGS by using 2D surfels to approximate thin surfaces. However, it falls short when dealing with glossy surfaces, resulting in visible holes in these areas. We find that the reflection discontinuity causes the issue. To fit the jump from diffuse to specular reflection at different viewing angles, depth bias is introduced in the optimized Gaussian primitives. To address that, we first replace the depth distortion loss in 2DGS with a novel depth convergence loss, which imposes a strong constraint on depth continuity. Then, we rectify the depth criterion in determining the actual surface, which fully accounts for all the intersecting Gaussians along the ray. Qualitative and quantitative evaluations across various datasets reveal that our method significantly improves reconstruction quality, with more complete and accurate surfaces than 2DGS. Code is available at \url{https://github.com/XiaoXinyyx/Unbiased_Surfel}.


\end{abstract}  
\section{Introduction}
\label{sec:intro}
During the past years, differentiable multi-view 3D reconstruction approaches~\cite{yao2018mvsnet,gu2019cas,mildenhall2021nerf,wang2021neus,barron2021mip,mueller2022instant} have rapidly emerged along with the neural network revolution. Among these, 3D Gaussian Splatting (3DGS)~\cite{kerbl20233d} is recognized as one of the most appealing methods. It represents scenes with a collection of 3D Gaussian primitives, achieving remarkable performance in real-time {photorealistic} novel view synthesis. Since this innovation, follow-up works have flourished in a short period, advancing 3DGS in such as rendering~\cite{malarz2023gaussian,lu2024scaffold,yu2024gaussian}, dynamic or large-scale scenes~\cite{wu20244d,liang2023gaufre,guo2024motion,lin2024vastgaussian,Hierarchical_3DGS}, and various other applications~\cite{deng2024compact,xu2024texture,yang2024gaussianobject}. 

However, direct geometry extraction using the resulting Gaussian centers only leads to inaccurate reconstruction due to the volumetric property of 3D Gaussians, which conflicts with the thin surface nature. To address this issue, SuGaR~\cite{guedon2023sugar} aligns Gaussians to surfaces via regularization, using depth maps to extract surface points. Other works~\cite{chen2023neusg,chen2024pgsr} attempt to flatten the 3D Gaussians during optimization. Recently, GOF~\cite{yu2024gaussian} proposed an opacity field to derive the isosurface. Nevertheless, 3D Gaussians still retain thickness. 

The recent work 2D Gaussian Splatting (2DGS)~\cite{huang20242d} makes a breakthrough, where 2D surfels are used instead of 3D Gaussians. 2D disks can align well with complex thin surfaces. Meanwhile, using 2D splats overcomes the inconsistency issue of 3D Gaussian primitives in multi-view projection, thereby significantly improving the geometric accuracy in reconstruction. Although theoretically sound, surface holes are often observed in 2DGS results, which arouses our curiosity; see, \eg, the 2nd column of Fig.~\ref{fig:teaser}. %

Upon closer inspection, we found almost all holes are located in highlight spots of glossy or semi-glossy areas. We attribute them to the discontinuous surface reflection under different viewing angles. Glossy surfaces manifest diffuse properties at most viewing angles, yet tend to overexposure at certain angles. To fit such an abrupt transition, the optimization has to create highlight Gaussians behind the real surface, so that they can be seen from the frontal angle while invisible from other oblique directions; see the illustration in Fig.~\ref{fig:illustration}. Although there is a depth distortion loss in 2DGS, the use of transmittance weights makes it insufficient to enforce depth continuity, leading to holes in these regions. 

In addition to the aforementioned issue, we found that the depth used to represent the object surface in 2DGS is also biased. In the original method, the actual surface is considered to be the median point of the Ray-Gaussian intersections, which is determined by the accumulated opacity when it reaches 0.5. For glossy areas, however, even if some Gaussians are correctly placed on the real surface, they will be optimized to have small opacity so that the volumetric rendering is correct along different viewing directions.

In this paper, we propose an unbiased 2D Gaussian Splatting to address the discontinuity issue and depth bias mentioned above, with two key components. First, we propose a \emph{depth convergence loss}, which forces the Gaussian depth to be continuous and smooth. Given the reliable depth around the highlight regions, the proposed constraint will promote an ``edge-growing'' effect from the specular-diffuse boundary to recover complete surface geometry for these areas. 
However, the highlight Gaussians will remain behind the surface using only the convergence loss. Therefore, we introduce \emph{depth correction}, a new criterion to identify the actual surface, which considers both the number of intersecting Gaussians and the accumulated opacity along the ray. With unbiased depth, our improved 2DGS substantially enhances reconstruction quality. Experiments on various datasets show that our method significantly outperforms the original 2DGS in terms of geometric accuracy.

\yyx{In summary, based on experimental analysis of glossy regions, we present the following contributions:} 
\begin{itemize}
    \item We propose a novel depth convergence loss \yyx{to enhance stronger geometric constraints, thereby promoting ``edge-growing'' for high-specular regions, leading to more complete and smoother surface reconstruction than the original 2DGS.}
    \item We introduce unbiased depth to represent the actual surface, further enhancing overall reconstruction accuracy.
\end{itemize}

\section{Related work}
\label{sec:related_work}
Recovering 3D geometry from 2D images is a fundamental yet challenging task, including both traditional methods [1, 14, 33] and deep-learning-based methods [8, 15, 38]. Recent advances in neural radiance fields bring new insights to multi-view-based 3D reconstruction. This section mainly reviews recent work on neural implicit surface reconstruction and Gaussian splatting methods.



\noindent\textbf{Neural Implicit Methods.} Neural radiance fields (NeRF)~\cite{mildenhall2021nerf} have achieved remarkable results in novel view synthesis (NVS) and have opened up new avenues for multi-view reconstruction. NeuS~\cite{wang2021neus} and VolSDF~\cite{Yariv:2021:Volume} map volume density to signed distance fields (SDF), allowing mesh extraction at any resolution using the Marching Cubes~\cite{lorensen1998marching}. NeuralWarp~\cite{darmon2022improving} enhances photometric consistency through a warping loss, improving reconstruction quality. Subsequent works~\cite{Fu2022GeoNeus,yu2022monosdf} leverage geometry priors, like normal priors and depth priors, have further refined reconstruction accuracy. Neuralangelo~\cite{li2023neuralangelo} employs hash encodings~\cite{mueller2022instant}, a more expressive scene representation, achieving highly detailed reconstructions. Although these methods yield promising surface reconstruction, the long training time remains a significant bottleneck. 

\noindent\textbf{Gaussian Splatting.} 
Recently, 3D Gaussian Splatting (3DGS)~\cite{kerbl20233d} has become one of the most popular 3D reconstruction methods, as it achieves high-quality real-time photorealistic NVS results. Follow-up works apply or improve 3DGS in various aspects, including rendering~\cite{malarz2023gaussian,lu2024scaffold,yu2024gaussian}, \yyx{exploring new shading methodologies for specular surfaces ~\cite{zhang2024refgs, jiang2023gaussianshader, ye2024gsdr, Tang3iGS}}, reconstructing dynamic or large-scale scenes~\cite{wu20244d,liang2023gaufre,guo2024motion,lin2024vastgaussian,Hierarchical_3DGS}, and various other applications~\cite{deng2024compact,xu2024texture,yang2024gaussianobject}. 
However, a remaining challenge is to derive high-accuracy geometry from the optimized Gaussian primitives.

SuGaR~\cite{guedon2023sugar} first addressed the geometry problem of 3DGS by regularizing the Gaussians to conform to surfaces. However, due to the inherent thickness of Gaussians, the extracted surface points are often noisy. Some recent works~\cite{chen2023neusg,lyu20243dgsr,yu2024gsdf,zhang2024neural} turn to SDF and jointly optimize 3DGS and neural SDF for surface reconstruction. GOF~\cite{yu2024gaussian} constructs a Gaussian opacity field upon 3D Gaussians, identifying level sets to determine surfaces. Spiking GS~\cite{zhang2024spiking} uses a minimal number of Gaussians, striking a good balance between geometric quality and cost.
To approximate the thin nature of surfaces, NeuSG~\cite{chen2023neusg} proposes to minimize the minimal axis of 3D Gaussians to be flat shapes. PSGR~\cite{chen2024pgsr} further incorporates multi-view geometric consistency, reaching remarkable reconstruction quality.  Nonetheless, the planar Gaussians adopted in these methods are still 3D primitives with thickness. 

To acquire real 2D splats, Gaussian Surfels~\cite{dai2024high} directly zeroes the third axis of 3D Gaussians, with other components of 3DGS unchanged. 2DGS~\cite{huang20242d} rebuilt the Gaussian Splatting system using pure 2D disks, which completely solves the inconsistency of multi-view projections brought by 3D Gaussians, achieving very high reconstruction quality. However, 2DGS encounters depth bias in glossy areas, resulting in holes or limited geometric details. Our method addresses 2DGS's limitations and introduces unbiased depth for surface reconstruction, significantly improving the geometry quality. 


\section{Method}
\label{sec:method}

\subsection{Preliminaries}
\textbf{3D Gaussian Splatting} represents 3D scenes with a set of 3D Gaussian primitives~\cite{kerbl20233d}. Each primitive is parameterized via a center point ${\bp}_{k}$ and a covariance matrix $\bSigma$:
\begin{equation}\label{...}
\mathcal{G}(\bp) = \exp(-\frac{1}{2}(\bp-\bp_{k})^{\top}\bSigma ^{-1}(\bp-\bp_{k} ))
\end{equation}
{where $\bSigma = \bR\bS\bS^{\top}\bR^{\top}$, with $\bR$ representing the rotation matrix and $\bS$ as the scaling matrix.} 3D Gaussians are transformed from the world coordinates to the camera coordinates using the transformation matrix $\bW$. An affine transformation $\bJ$~\cite{zwicker2001ewa} is applied to the projection to the image plane, resulting in the transformed covariance matrix: 
\begin{equation}\label{...}
\bSigma ^{'} =\bJ \bW\bSigma \bW^{\top} \bJ^{\top}
\end{equation}

The first two rows and columns of $\bSigma'$ are extracted to derive a 2D Gaussian value $\mathcal{G}^{2D}(\bx)$, where $\bx=(x,y)$ denotes an image coordinate. 3DGS~\cite{kerbl20233d} employs $\mathcal{G}^{2D}(\bx)$ for front-to-back volumetric alpha blending:
\begin{equation}\label{...}
\bc(\bx) = \sum_{i=1}^{n} \bc_{i} \alpha _{i}\mathcal{G}^{2D}_{i} (\bx)\prod_{j=1}^{i-1}(1-\alpha _{j}\mathcal{G}^{2D}_{j}(\bx))
\end{equation}
where $\alpha_{i}$ represents the alpha blending weight and $\bc_{i}$ is the view-dependent color associated with the $i$-th Gaussian.

\noindent\textbf{2D Gaussian Splatting} %
uses 2D surfels to represent 3D scenes~\cite{huang20242d}. A 2D splat is composed of a center $\bp_k$, two principal tangential vectors $\bt_u$ and $\bt_v$, and a scaling vector $\bS = (s_u, s_v)$. By computing the ray-splat intersection $\bu(\bx)$ in $uv$ space and applying an object-space low-pass filter~\cite{botsch2005high}, its 2D Gaussian value can be evaluated by:
\begin{equation}\label{...}
\mathcal{G}(\bu(\bx)) = \exp(-\frac{u^{2} + v^{2}  }{2} )  
\end{equation}
\begin{equation}\label{...}
\mathcal{\hat{G}}(\bx)   = \max\left\{\mathcal{G}(\bu(\bx)),\mathcal{G}(\frac{\bx-\bc}{\sigma }) \right\}
\end{equation}
where $\bc$ denotes the image coordinate of the projection of $\bp_k$, and $\sigma$ is the radius of screen-space low-pass filter. $\sigma$ is set to $\sqrt{2}/2$ to handle degenerate cases when 2D Gaussian is rasterized from a slanted viewing angle.

2DGS uses a rasterization process similar to that of 3DGS~\cite{kerbl20233d}. Initially, screen-space bounding boxes are computed for each Gaussian primitive. Then, 2D Gaussians are sorted by depth and organized into tiles based on their bounding boxes. Finally, volumetric alpha blending is performed from front to back until the accumulated opacity reaches saturation, {which can be formulated as:} 
\begin{equation}\label{...}
\bc(\bx) = \sum_{i=1}^{n} \bc_{i} \alpha _{i}\hat{\cG_i} (\bx)\prod_{j=1}^{i-1}(1-\alpha _{j}\hat{\cG_j}(\bx)) 
\end{equation}

\subsection{Hole Formation in Glossy Areas}
\label{hole formation sec}
Based on our observations, in regions where specular highlights appear on glossy surfaces, only a few Gaussian surfels can adhere to the actual surface in the 2DGS results. In contrast, most Gaussian surfels tend to concave inward in these areas, resulting in holes or pits in the reconstruction. 

\begin{figure}
    \centering
    \includegraphics[width=1.0\linewidth]{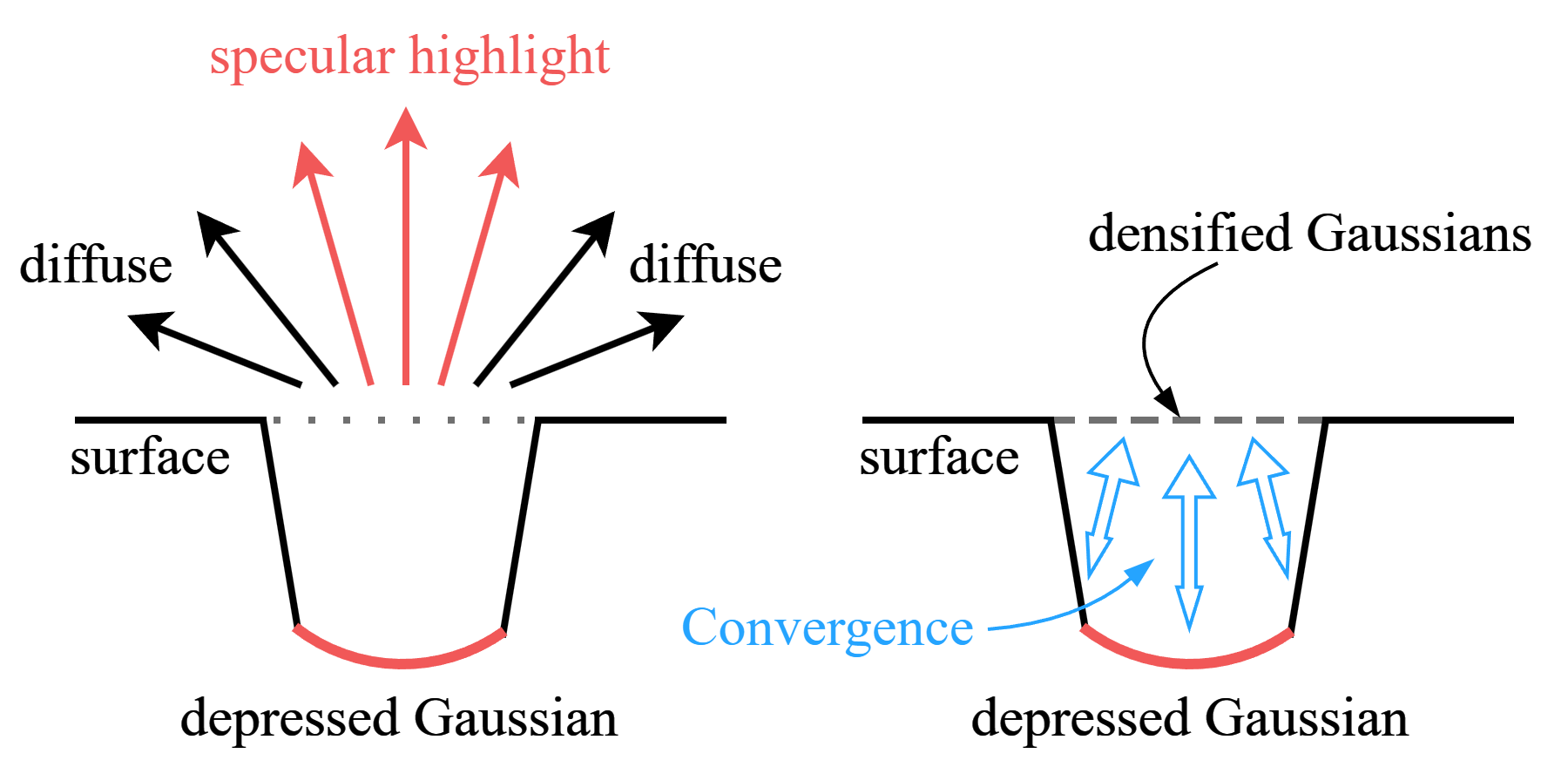}
    \caption{Illustration of the depressed Gaussians and {the proposed} depth convergence process. 2DGS encounters challenges in accurately reconstructing glossy surfaces, while our depth convergence loss can help overcome this shortcoming.}
    \label{fig:illustration}
\end{figure}

As illustrated in Fig.~\ref{fig:illustration}, we analyzed the reasons for this phenomenon. Glossy or semi-glossy surfaces with high smoothness tend to reflect extremely high radiance in specific viewing directions, such as when the viewing direction and the light source direction are symmetric about the surface normal. In most other viewing directions, the radiance does not change significantly, similar to diffuse surfaces. The spherical harmonics of Gaussian surfels alone cannot fit the outgoing radiance that changes dramatically with the viewing direction. Therefore, when dealing with such surfaces, the optimization process will put some concave Gaussian splats primarily responsible for replicating the object's highlight. As a result, these Gaussians can only be captured by the camera at specific viewing angles. For most other angles, only non-highlight Gaussians are visible. The system cleverly utilizes the occlusion relationship of the Gaussians to compensate for the inability of spherical harmonics to effectively replicate high-frequency information.

\begin{figure}[t]
    \centering
    \includegraphics[width=1.0\linewidth]{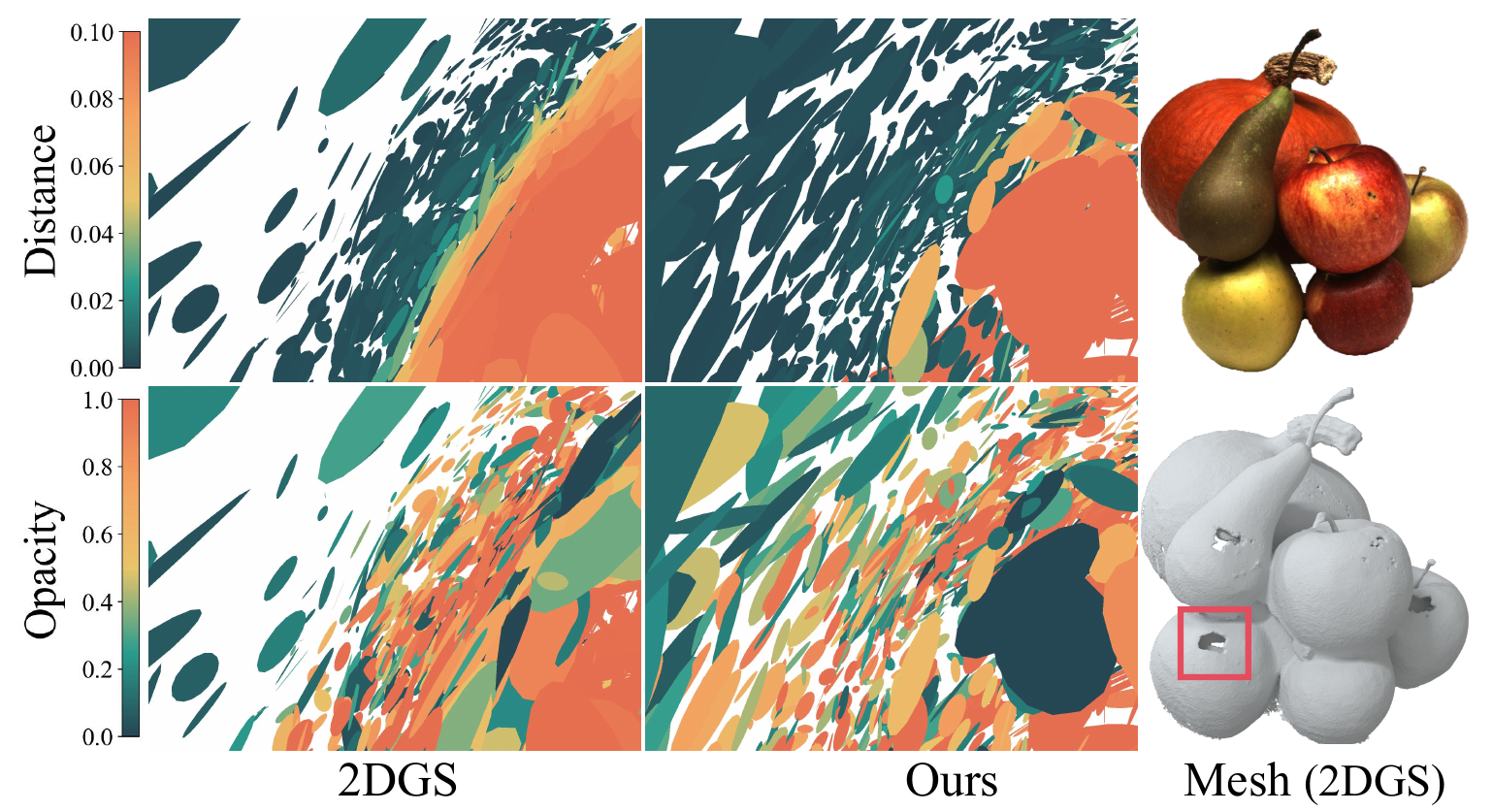}
    \caption{ Visualization of the resulting 2D Gaussian primitives in a high-specular region. The top row shows the distance of Gaussian centers to the ground truth point cloud. The bottom row visualizes the opacity of Gaussians. We can see only a few Gaussians with low opacity located on the real surface in the 2DGS result, while our method successfully densifies the correct surfels and converges the nearby Gaussians to the surface, leading to more continuous and complete reconstruction.}
    \label{fig:vis_gs}
\end{figure}

Fig.~\ref{fig:vis_gs} visualizes a real case of the resulting Gaussians from 2DGS. It clearly shows that in glossy areas, only a few transparent (low-opacity) Gaussian primitives are left at the actual surface, whereas most splats with much greater opacity concave inward after the optimization. We further attribute the issue to the depth distortion loss proposed in 2DGS~\cite{huang20242d}, 
where the intersected primitives along a ray are constrained to be close to each other using this loss. However, the multiplication of opacity as weights makes those splats with higher opacity dominate the loss calculation (see Eq.~13 in their paper). Therefore, in regions of specular highlights, it will lead to the clustering of Gaussian surfels towards the inward-concave, forming a ``false'' surface. When extracting the object mesh, the algorithm fetches biased depth in these areas, resulting in holes or pits. 

\subsection{Depth Convergence Loss}
As discussed, in the glossy regions of the object, the surface contains a sparse distribution of low-opacity Gaussians. To improve Gaussian coveragence on the actual surface without adversely impacting other diffuse surfaces, we encourage Gaussians to move toward the actual surface during optimization, thereby enhancing reconstruction quality. Accordingly, we introduce a new depth convergence loss, and use it to replace the depth distortion loss in 2DGS~\cite{huang20242d}.

In the rendering pipeline of tile-based Gaussian Splatting, the Gaussians are depth-sorted by their center positions. For each ray emitted from the camera, a depth array with an approximately ordered sequence can be obtained, where each element represents the depth value of the ray-splat intersection. Specifically, we denote the depth value of the $i$-th intersection as $d_i$, and the squared depth difference between two adjacent intersections as $D_i = (d_i-d_{i-1})^2$. The depth convergence loss is defined as follows:
\begin{equation}\label{...}
\mathcal{L}_{\text{converge}} = \sum_{i = 2}^{n}\min(\hat{\cG_i}(\bx),\hat{\cG}_{i-1}(\bx))D_{i} 
\end{equation}
where $\hat{\cG}(\bx_{i} )$ represents the 2D Gaussian value of the $i$-th Gaussian, {which is only} used as the weight for $D_i$ during optimization and does not participate in gradient computation. 
As analyzed previously, we omit the opacity to treat Gaussians with low opacity equally. Our loss functions like a Dirichlet energy minimization, wherein object surfaces—characterized as physical discontinuities along light transport paths—naturally constitute global minima in the energy space. If the Gaussian primitives converge but are not aligned to the surface, it would inevitably produce view-dependent artifacts, triggering further optimization. Consequently, although the proposed depth convergence loss operates only on the spatial properties of Gaussian splats to improve their continuity, it will also drive their alignment with the underlying geometric surface.

To further encourage concave Gaussians to converge toward the actual surface, we scale the partial derivative of $D_i$ w.r.t. $d_i$ by a factor greater than 1 during gradient calculation: $\frac{\partial}{\partial d_i}D_i \leftarrow 2k(d_i-d_{i-1})$, where $k=1.25$. \yyx{This modification encourages Gaussians to shift toward the camera direction. For concave highlight regions where multi-view consistency constraints prove less effective, our loss function progressively reduces concavity depth. In strongly textured regions, multi-view consistency constraints dominate the optimization, while the cconvergence loss primarily serves to aggregate Gaussians.}

\begin{figure}
    \centering
    \includegraphics[width=1.0\linewidth]{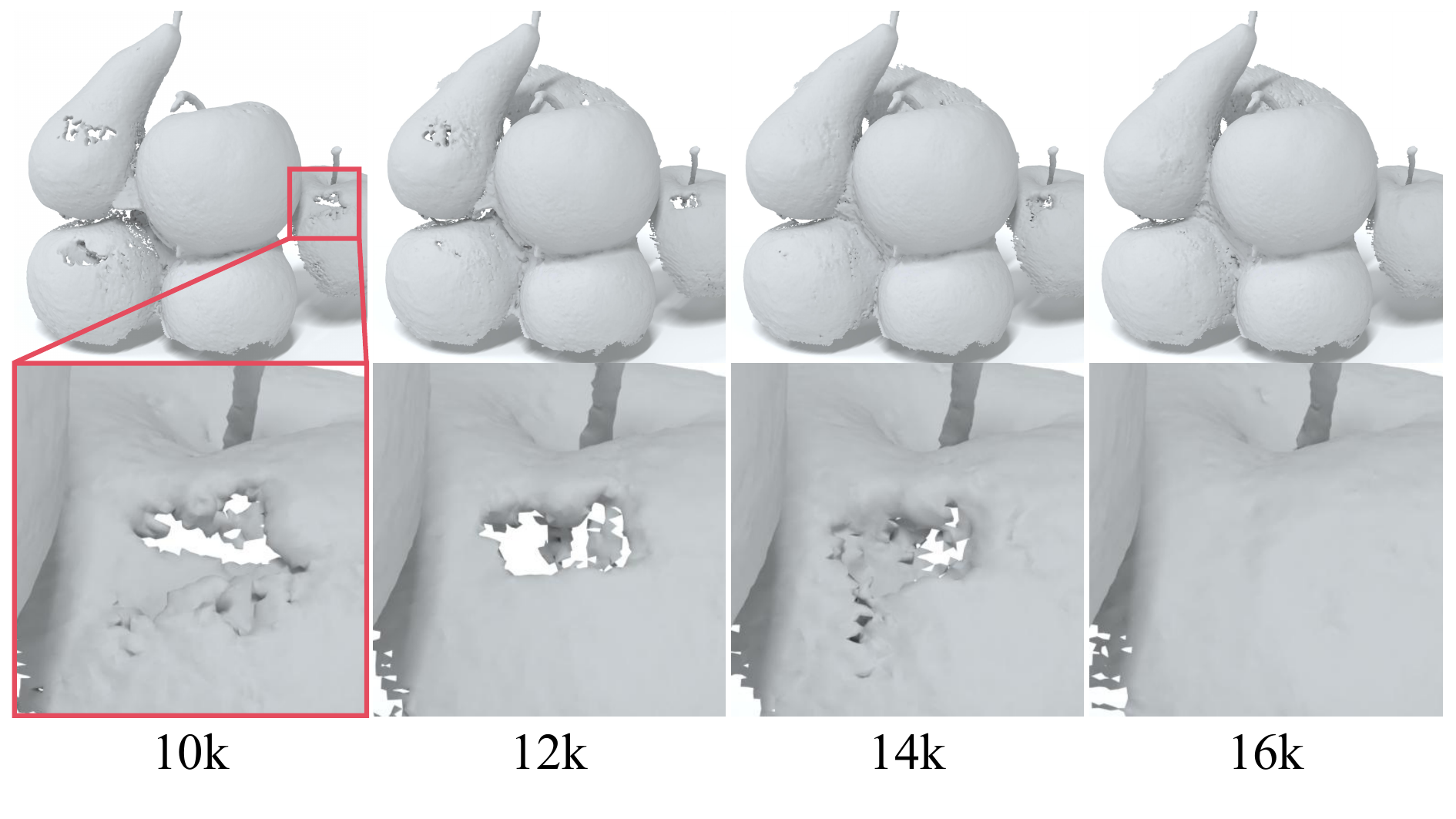}
    \caption{{The ``edge-growing'' process brought by our depth convergence loss} at different optimization iterations.}
    \label{fig:edge_grow}
\end{figure}

Fig.~\ref{fig:edge_grow} visualizes the process of hole filling brought by our depth convergence loss. In regions with holes, the loss gradually pulls concave Gaussians from the hole margins to the true surface, exhibiting an ``edge-growing'' phenomenon along with optimization. Our loss also induces larger gradients for Gaussians in glossy areas, which might further trigger the Gaussian splitting and cloning mechanism~\cite{kerbl20233d}, thereby generating additional Gaussians to fill the holes.

\subsection{Depth Correction}
2DGS considers the ray-splat intersection where the accumulated opacity reaches 0.5 as the actual surface. In implementation, the accumulated transmittance is empirically adopted:
\begin{equation}\label{eq:transmittance}
T_{i} =\prod_{j=1}^{i-1}(1-\alpha _{j}\hat{\cG_j}(\bx))
\end{equation}
When $T_{i}$ decreases to 0.5, nearby depth points are employed to compute the surface normal with finite differences~\cite{huang20242d}.

Using the cumulative product of transmittance has an obvious disadvantage. As shown in Fig.~\ref{fig:vis_gs}, for specular highlight regions on glossy surfaces, there are quite dense Gaussian splats located on the surface after optimization using our depth convergence loss, whereas most of them are still with low opacity. Therefore, when using the depth criterion above, the accumulated transmittance is usually insufficient
for the algorithm to capture the actual surface at these areas. 

Based on this, we propose to consider both the number of Gaussians involved in alpha blending and the opacity along the ray to determine the surface position. Specifically, we define a cumulative opacity $O_i$ in the form of summation: 
\begin{equation}\label{eq:opacity}
O_{i}  = \sum_{j=1}^{i} (\alpha _{j}+ \epsilon) \hat{\cG}_j(\bx) 
\end{equation}
where $\epsilon$ is a small constant compensating opacity, and can be used to control the impact of the cumulative number of intersected Gaussians on $O_i$. We set $\epsilon$ as 0.1 in all our experiments. When the accumulated opacity exceeds a threshold (0.6 according to experiments), the intersection point $\bp_s$ is considered as the actual surface point of the object. See our supplementary material for more analysis of depth correction.

Next, we derive the normal map from the gradients of the rectified depth map and align it with the splats' normal~\cite{huang20242d}:

\begin{equation}\label{...}
\bN(x,y)=\frac{\nabla_{x}\bp_s  \times \nabla_{y}\bp_s }{\left |\nabla_{x}\bp_s  \times \nabla _{y}\bp_s  \right | }
\end{equation}
\begin{equation}\label{...}
\cL_n = \sum_{i}^{}\omega_{i}(1 - \bn_{i}^{\top}\bN ) 
\end{equation}
where $\bN$ is the normal estimated {at the surface point}, $\bn_{i}$ represents the normal of the $i$-th splat that is oriented towards the camera, and $w_{i} $ denotes the blending weight of the $i$-th intersetion point. 

\subsection{Final Loss}
Following 3DGS~\cite{kerbl20233d}, we combine $\cL_{1}$ loss with D-SSIM as the color reconstruction loss. Inspired by PGSR~\cite{chen2024pgsr}, we applied gamma correction with a gamma value of $0.5$ to the ground truth images to enhance contrast in dark areas, and then calculated the color loss with the rendered image:
\begin{equation}
\cL_c=(1-\lambda)\left | I_{i}^{r} - \sqrt{I_{i}^{g}}  \right |_{1}+\lambda \cdot SSIM(I_{i}^{r},\sqrt{I_{i}^{g}} )    
\end{equation}
where subscript $i$ indexes the camera view, $I_i^{r}$ is the differentiable rendered image, $I_i^{g}$ represents the ground truth image, {$\sqrt{\cdot}$ denotes gamma correction}, and $\lambda$ weights the importance of the D-SSIM term.
The total loss is defined as:
\begin{equation}\label{...}
\cL = \cL_c+\lambda_n \cL_n+\lambda_c  \cL_{\text{converge}}
\end{equation}
We set $\lambda=0.2$, $\lambda_{n}=0.05$, and $\lambda_c=7$ in all experiments. 

\section{Experiments}
\label{sec:exp}


\subsection{Experimental Setup}

\noindent\textbf{Implementation Details.} Building upon the 2DGS framework, we modify its CUDA kernel to output depth convergence and depth correction maps. We make a slight change to the adaptive Gaussian densification strategy, setting it to stop at 20,000 iterations. After stopping densification, we clamp the scale of Gaussians so that they do not exceed the threshold set during densification. Once the depth difference between two adjacent ray-splat intersections exceeds a threshold (set to one-fourth of the scene radius), we exclude the depth convergence loss calculation between them. In addition, we only apply gamma correction on the DTU dataset~\cite{jensen2014large}. We conduct all the experiments on a single NVIDIA RTX4090 GPU with 24GBytes memory.

\noindent\textbf{Datasets.} We evaluate our method on three datasets, including DTU~\cite{jensen2014large}, Tanks and Temples~\cite{knapitsch2017tanks}, and Mip-NeRF360~\cite{barron2022mip}. The DTU dataset is a widely used benchmark for evaluating object-level 3D reconstruction quality. Each scene in the dataset contains 49 or 64 images, but does not fully cover all viewpoints of the object. {Following the previous work}~\cite{huang20242d,yu2024gaussian}, we selected 15 scenes for comparison, which include both highlights and shadows, making them more challenging for reconstruction. We use Chamfer distance as the geometry metric, which measures the mean of reconstruction completeness and accuracy. 

\noindent\textbf{Baselines.} Following~\cite{huang20242d, yu2024gaussian}, we compare our method with state-of-the-art neural implicit surface methods, including VolSDF~\cite{Yariv:2021:Volume}, NeuS~\cite{wang2021neus}, and Neuralangelo~\cite{li2023neuralangelo}. For Gaussian splatting methods, we compare with 3DGS~\cite{kerbl20233d}, SuGaR~\cite{guedon2023sugar}, Gaussian Surfels (G-Surfel)~\cite{dai2024high}, 2DGS~\cite{huang20242d}, and GOF~\cite{yu2024gaussian}. 

\setlength\tabcolsep{0.5em}
\begin{table*}[t]
\caption{Quantitative results of chamfer distance (mm)$\downarrow$ on DTU dataset~\cite{jensen2014large}. Our method achieves the highest geometry accuracy compared to other Gaussian splatting methods. {It's also worth noting that our method surpasses the original 2DGS on all scenes.}} 
\centering
\resizebox{\textwidth}{!}{%
\begin{tabular}{@{}lccccccccccccccccclc|c}
\toprule
 \multicolumn{3}{c}{} & 24 & 37 & 40 & 55 & 63 & 65 & 69 & 83 & 97 & 105 & 106 & 110 & 114 & 118 & 122 & & Mean & Time \\ \midrule
 & VolSDF~\cite{Yariv:2021:Volume} & & 1.14 & 1.26 & 0.81 & 0.49 & 1.25 & 0.70 & \tbest0.72 & \sbest1.29 & 1.18 & \sbest0.70 & 0.66 & 1.08 & 0.42 & 0.61 & 0.55 & & 0.86 & \texttt{>} 12h\\
 & NeuS~\cite{wang2021neus} & & 1.00 & 1.37 & 0.93 & 0.43 & 1.10 & \sbest 0.65 &  \sbest0.57 & 1.48 & \tbest1.09 & 0.83 & \tbest0.52 & 1.20 & \sbest0.35 & \tbest0.49 & 0.54 & & 0.84 & \texttt{>} 12h\\
 
 & Neuralangelo~\cite{li2023neuralangelo} & & \best 0.37 & \sbest0.72 & \sbest 0.35 & \best 0.35 & \sbest 0.87 & \best 0.54 & \best 0.53 & \sbest1.29 & \best 0.97 & \tbest0.73 & \best 0.47 & \sbest 0.74 & \best 0.32 & \best 0.41 &  \sbest 0.43 & & \best 0.61 & \texttt{>} 128h\\ \midrule

 & 3DGS~\cite{kerbl20233d} && 2.14 & 1.53 & 2.08 & 1.68 & 3.49 & 2.21 & 1.43 & 2.07 & 2.22 & 1.75 & 1.79 & 2.55 & 1.53 & 1.52 & 1.50 & & 1.96 & \tbest8.1m\\
 
 & SuGaR~\cite{guedon2023sugar} && 1.47 & 1.33 & 1.13 & 0.61 & 2.25 & 1.71 & 1.15 & 1.63 & 1.62 & 1.07 & 0.79 & 2.45 & 0.98 & 0.88 & 0.79 & & 1.33 & $\sim$ 1h\\

 & G-Surfel~\cite{dai2024high}  && 0.66 & 0.93 & 0.54 & 0.41 & 1.06 & 1.14 & 0.85 & \sbest1.29 & 1.53 & 0.79 & 0.82 & 1.58 & 0.45 & 0.66 & 0.53 & & 0.88 & \sbest7.9m\\
 
 & 2DGS~\cite{huang20242d} && \tbest 0.48 & 0.91 & 0.39 & \tbest0.39 & \tbest1.01 & 0.83 & 0.81 & 1.36 & 1.27 & 0.76 & 0.70 & 1.40 & 0.40 & 0.76 & 0.52 & & 0.80 &  \best7.5m\\
 
  & GOF~\cite{yu2024gaussian} & &  0.50 & \tbest 0.82 &  \tbest0.37 &  \sbest0.37 & 1.12 &  0.74 & 0.73 &  \best 1.18 & 1.29 & \best 0.68 & 0.77 &  \tbest0.90 & 0.42 & 0.66 &  \tbest0.49 &&  \tbest0.74 & 23m\\
  
  & Ours && \sbest 0.43 & \best 0.71 & \best 0.32 & \best 0.35 & \best 0.84 & \tbest 0.66 & \sbest 0.57 &  \tbest1.34 & \sbest 1.06 & \tbest 0.73 & \sbest 0.48 & \best 0.72 & \tbest 0.36 & \sbest 0.44 & \best 0.42 & & \sbest 0.63 & 9.5m\\
 \bottomrule
\end{tabular}
}
\label{tab:dtu_result}
\end{table*}
\begin{table}[htbp]
\vspace{-2pt}
\centering
\caption{Quantitative comparisons on Mip-NeRF360~\cite{barron2022mip}. All scores of the baseline methods are directly taken from their papers (whenever available).}
\resizebox{0.98\columnwidth}{!}{
\begin{tabular}{@{}l|ccc|ccc}
 & \multicolumn{3}{c@{}|}{Outdoor Scene} & \multicolumn{3}{c@{}}{Indoor scene} \\ 
& PSNR~$\uparrow$ & SSIM~$\uparrow$ & LIPPS~$\downarrow$ & PSNR~$\uparrow$ & 
SSIM~$\uparrow$ & LIPPS~$\downarrow$ \\
\hline
3DGS~\cite{kerbl20233d} & \sbest24.64 & \sbest0.731 & \sbest0.234 & \sbest30.41 & \sbest0.920 & 0.189 \\

2DGS~\cite{huang20242d} & 24.33 & 0.709 & 0.284  & \tbest30.39 & \best0.924 & \best0.182  \\
GOF~\cite{yu2024gaussian} & \best24.82 & \best0.750 & \best0.202 & \best30.79 & \best0.924 & \sbest0.184 \\
Ours & \tbest24.60 & \tbest0.728 & \tbest0.255 & 30.00 & \tbest0.918 & \tbest0.185 \\
\end{tabular}
}
\label{tab:mipnerf360}
\vspace{-2pt}
\end{table}
\begin{table}[htbp]
\caption{Ablation study of our method on DTU dataset~\cite{jensen2014large}.}
\centering
\vspace{-2pt}
\resizebox{\linewidth}{!}{
\begin{tabular}{@{}l|ccc}
 & Accuracy~$\downarrow$ & Completion ~$\downarrow$ & Average~$\downarrow$ \\
\hline
A. 2DGS~\cite{huang20242d} & 0.78 & 0.83 & 0.80\\
\hline
B. 2DGS + converge & 0.71 & 0.79 & 0.75 \\
C. 2DGS + converge + depth\_corr & 0.67 & 0.76 & 0.71 \\
\hline
D. Full Model & 0.59 & 0.67 & 0.63\\
\end{tabular}%
}
\label{tab:ablation}
\vspace{-5pt}
\end{table}

\begin{figure}[t]
    \centering
    \includegraphics[width=1.0\linewidth]{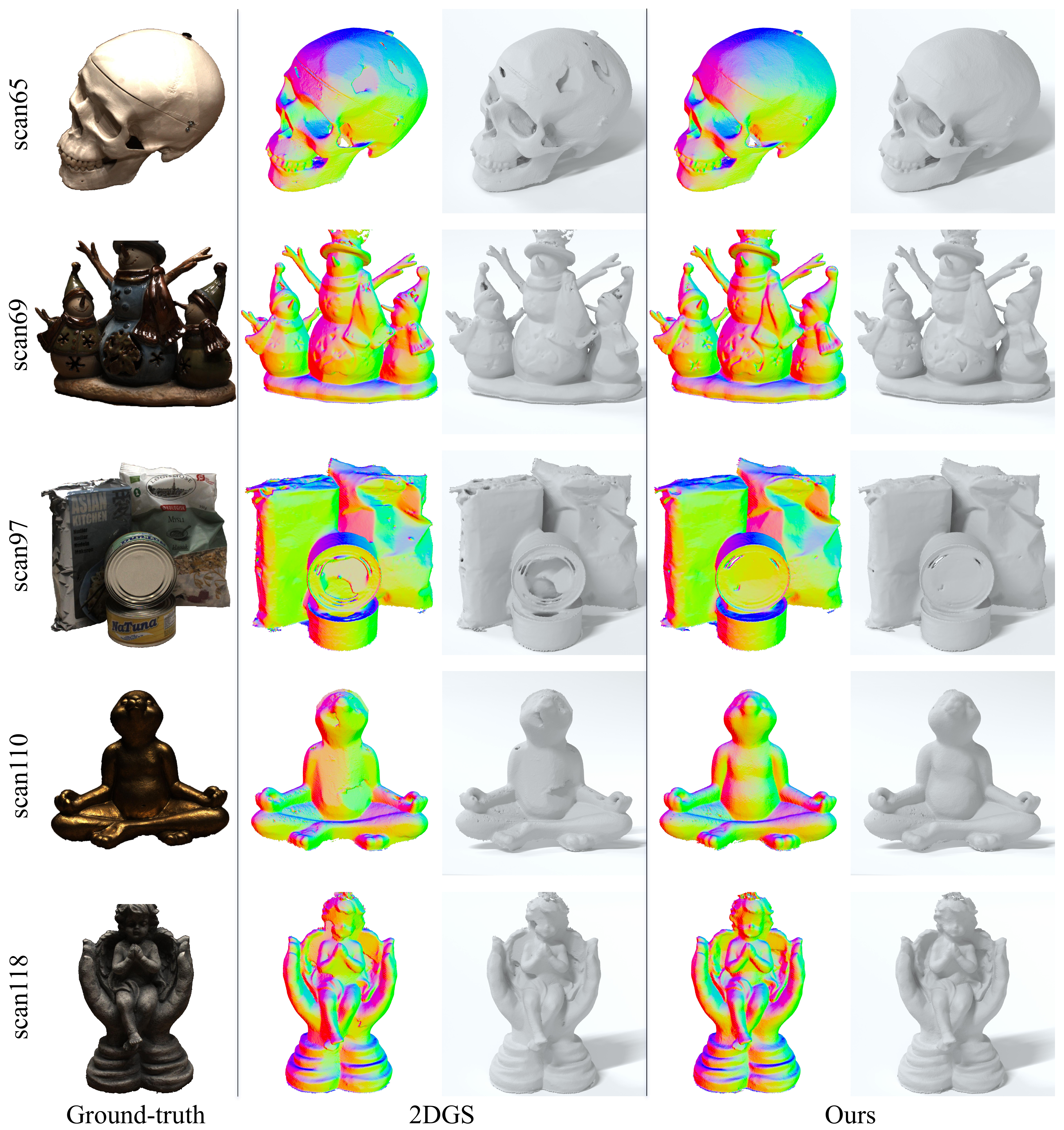}
    \caption{Qualitative comparison with 2DGS~\cite{huang20242d} on the DTU dataset~\cite{jensen2014large}. Following 2DGS, meshes are extracted by applying TSDF to depth maps. The ground truth images are shown at the leftmost, followed by the normals and extracted meshes. 
    Our method effectively fills holes in specular highlight areas and produces more complete and accurate surface reconstructions. A full qualitative comparison on DTU is included in the supplementary.}
    \label{fig:dtu_results}
\end{figure}

\begin{figure*}[htbp]
    \centering
    \includegraphics[width=1.0\linewidth]{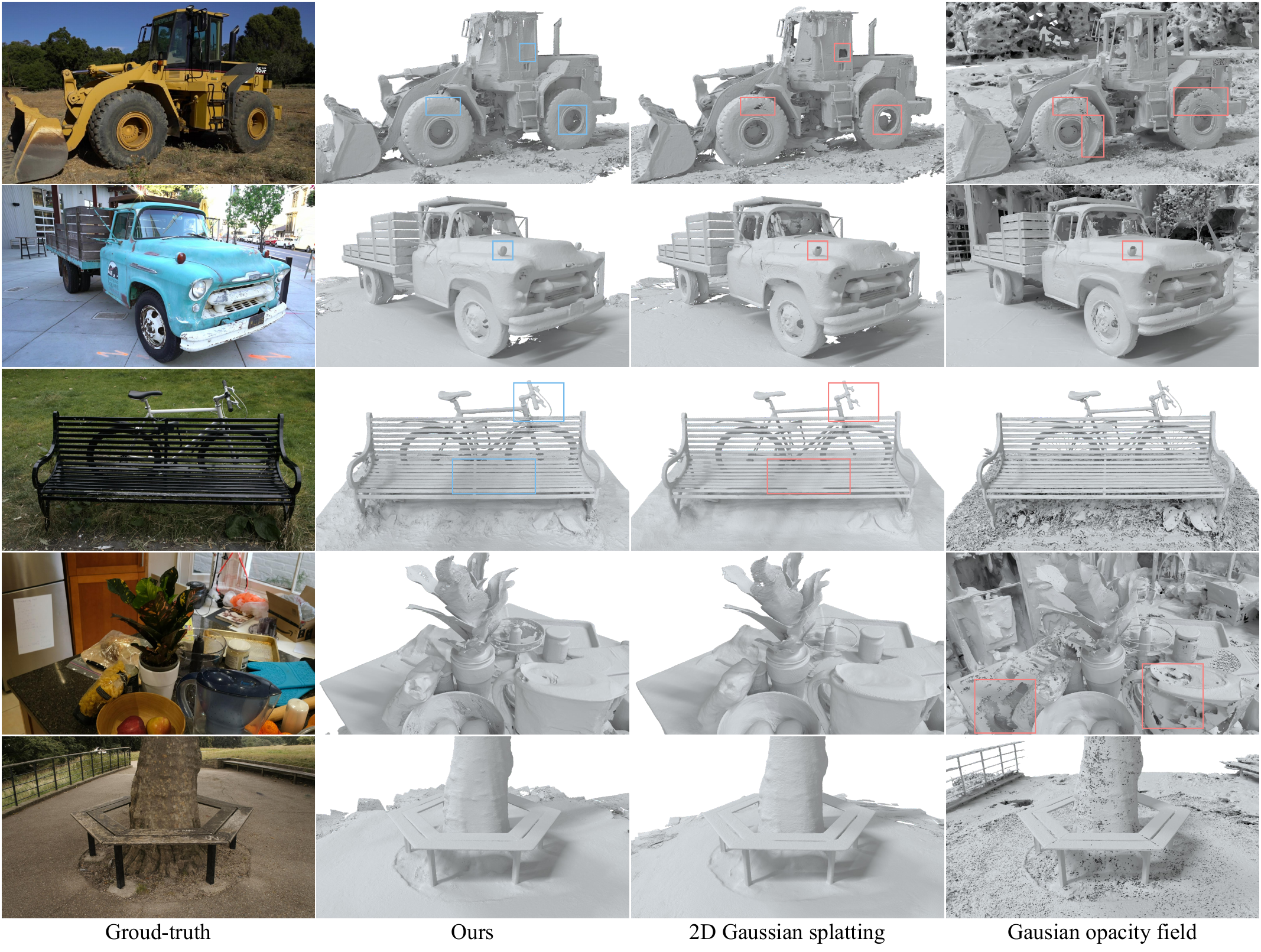}
    \caption{Qualitative comparisons with 2DGS~\cite{huang20242d} and GOF~\cite{yu2024gaussian}
    on Tanks \& Temples~\cite{knapitsch2017tanks} (the first 2 rows) and Mip-NeRF360~\cite{barron2022mip} (the last 3 rows). 
    Our method has fixed many artifacts caused by highlight reflections visible in the results of the other two methods. Besides, compared to GOF, a 3DGS-based method, 2DGS and our method produce smoother and more continuous surfaces.
    }
    \label{fig:Tnt&MipNerf}
\end{figure*}

\subsection{Geometry Reconstruction}
\label{subsec:geometry_quality}
We first compare the geometry quality of our method on the DTU dataset~\cite{jensen2014large} with all the baseline methods. 
As shown in Table~\ref{tab:dtu_result}, our method achieves the highest mean accuracy compared to other Gaussian splatting methods. In particular, our method consistently surpasses the original 2DGS in all scenes. Additionally, it attains comparable geometry reconstruction quality to the SOTA neural implicit methods, while offering a training speed improvement of over 100 times. Fig.~\ref{fig:dtu_results} further presents the qualitative comparisons with 2DGS on DTU, where the improvement of our method on the holes in glossy regions is clearly shown.

We then perform a qualitative comparison of reconstruction quality on Tanks and Temples~\cite{knapitsch2017tanks} and Mip-NeRF360~\cite{barron2022mip} with 2DGS~\cite{huang20242d} and GOF~\cite{yu2024gaussian}. As shown in Fig~\ref{fig:Tnt&MipNerf}, compared to the other two methods, our approach resolves many artifacts caused by specular reflections and excels in capturing sharp edges and intricate geometric details. Even when applied to outdoor scenes with larger dynamic ranges and more complex lighting, our method still performs well, fully demonstrating its robustness.



\subsection{Novel View Synthesis}
Although we build our method aiming at improving geometry quality, we still evaluate the quality of novel view synthesis and compare it to 3DGS~\cite{kerbl20233d}, 2DGS~\cite{huang20242d}, and GOF~\cite{yu2024gaussian} using the Mip-NeRF360 dataset~\cite{barron2022mip}. As listed in Table~\ref{tab:mipnerf360}, \yang{our method shows slightly better performance than the original 2DGS.} \yyx{The convergence constraint compels Gaussian kernels to distribute along object surfaces, which benefits geometric reconstruction but potentially compromises rendering quality. As demonstrated in section~\ref{hole formation sec}, the model exploits geometric misalignment to reconstruct object specular highlights. When such misalignment is eliminated, the smooth spherical harmonics (SH) representation struggles to represent high-frequency color variations. }

\begin{figure*}[htbp]
    \centering
    \includegraphics[width=0.97\linewidth]{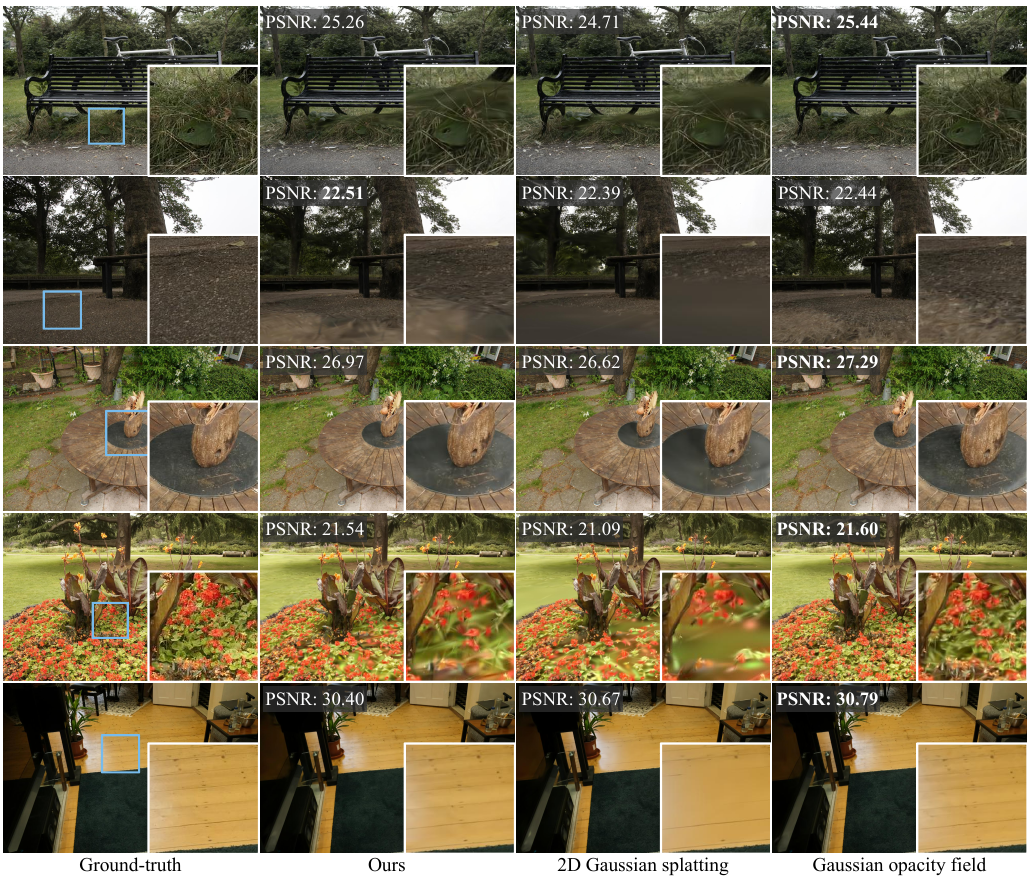}
    \caption{Qualitative comparisons of novel view synthesis with 2DGS~\cite{huang20242d} and GOF~\cite{yu2024gaussian} on Mip-NeRF360~\cite{barron2022mip}. 
    \yyx{
    Our renderings exhibit more high-frequency details and fewer artifacts than 2DGS for most planar parts, while falling behind in local dense and cluttered areas (rows 1 and 4).
    }}
    \label{fig:mipnerf360}   
\end{figure*}

Fig.~\ref{fig:mipnerf360} shows the qualitative comparisons of novel view synthesis on Mip-NeRF360 with 2DGS and GOF. \yyx{Compared with GOF, our method reaches comparable renderings in planar parts (rows 2, 3, and 5 in Fig.~\ref{fig:mipnerf360}), while falling behind in some local dense and tangled regions (rows 1 and 4 in Fig.~\ref{fig:mipnerf360}). We believe the advantage of GOF can be attributed to the use of 3D Gaussians, where small and dense 3D primitives are generated to fit these parts, leading to more visually appealing renderings.}

\begin{figure}[!htbp]
    \centering
    \includegraphics[width=1.0\linewidth]{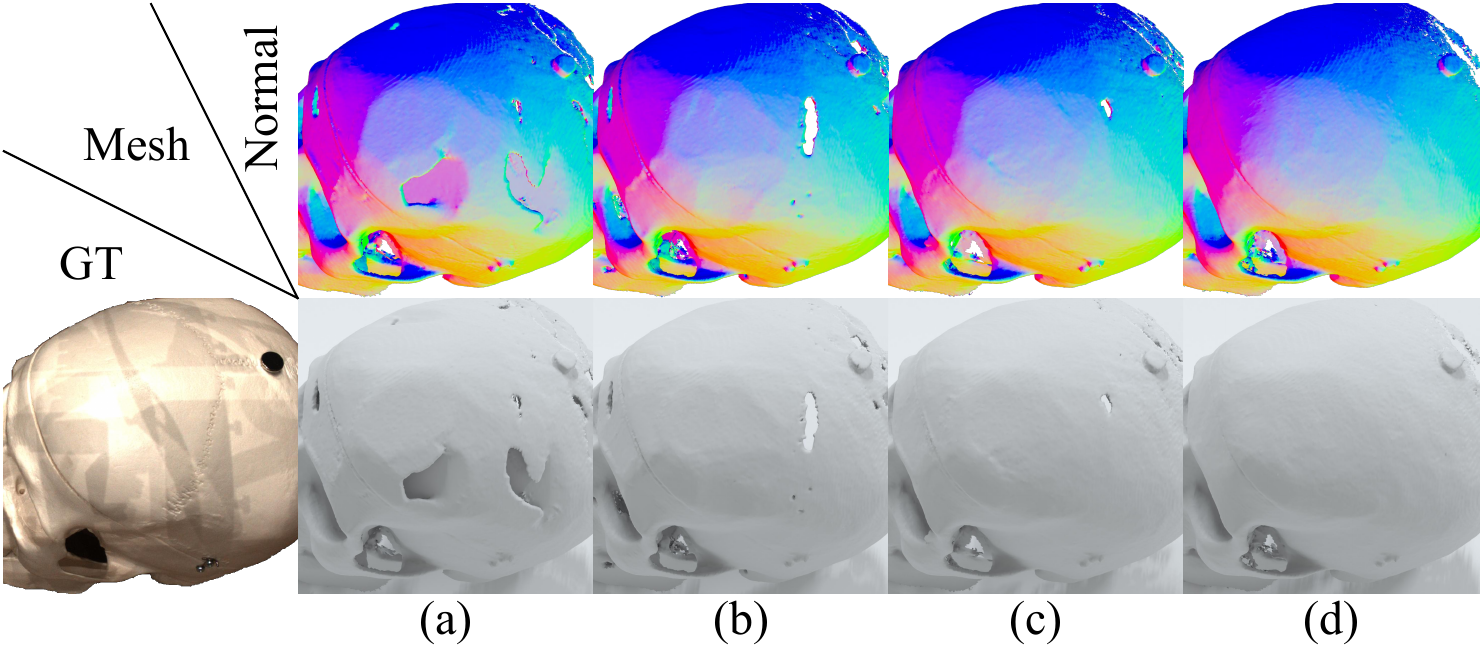}
    \caption{Ablation study: (a) 2DGS; (b) 2DGS + Convergence loss; (c) 2DGS + Convergence loss + depth correction; (d) Our full model (\ie, + gamma correction)}
    \label{fig:ablation}
\end{figure}

\subsection{Ablation Study}
Table~\ref{tab:ablation} and Fig.~\ref{fig:ablation} present the ablation study on DTU dataset~\cite{jensen2014large}, where we progressively integrate our techniques into 2DGS to evaluate their impact on geometry quality. As shown in Fig.~\ref{fig:ablation} (b), when only the depth convergence loss is applied (Table~\ref{tab:ablation} B), it only partially addresses the issue of holes in glossy areas. When both of them are applied (Table~\ref{tab:ablation} C), the holes on the surface are effectively resolved; see Fig.~\ref{fig:ablation} (c). The full model with gamma correction achieves the best, showing improved geometric details in shadowed areas, as illustrated in Fig.~\ref{fig:ablation} (d) and Table~\ref{tab:ablation} D.


\section{Conclusion}
\label{sec:conclusion}

We have carefully analyzed the issue of 2DGS in handling glossy surfaces where specular highlights occur, and proposed two countermeasures: a new loss that ensures depth continuity in these areas and a new criterion to determine the unbiased depth for the actual surface. We compare with previous methods in terms of geometry and rendering quality, demonstrating the effectiveness and superiority of our method comprehensively. A remaining limitation lies in representing local dense and tiny parts in complex scenes, such as the weed and the flower bush shown in Fig.~\ref{fig:mipnerf360}.


\FloatBarrier

\section*{Acknowledgments}
This work was supported in parts by NSFC (U21B2023), ICFCRT (W2441020), GD Basic and Applied Basic Research Foundation (2023B1515120026), DEGP Innovation Team (2022KCXTD025), SZU Teaching Reform Key Program (JG2024018), and Scientific Development Funds from Shenzhen University.

{
\bibliographystyle{eg-alpha-doi}  
\bibliography{main.bib}
}

\appendix


\section{Evaluation metric on geometric accuracy}
\yang{As the DTU dataset~\cite{jensen2014large} has ground truth surface point clouds, we use Chamfer distance to evaluate the geometric quality as previous methods~\cite{huang20242d, yu2024gaussian}.} 
Chamfer distance (CD) is a commonly used distance metric in fields such as point cloud registration and 3D reconstruction, which calculates the similarity between two point sets. Specifically, its calculation consists of two parts:
\begin{equation}\label{...}
d_{\text{CD}}(P, Q) = \frac{1}{|P|} \sum_{p \in P} \min_{q \in Q} \| p - q \|^2 + \frac{1}{|Q|} \sum_{q \in Q} \min_{p \in P} \| q - p \|^2,
\end{equation}
where $P$ and $Q$ are two point cloud sets, representing the input data point cloud and the reference point cloud, respectively. \yang{The two terms above are also termed as Accuracy and Completion~\cite{huang20242d, yu2024gaussian}, where Accuracy indicates the proximity of points in the reconstructed model to the ground truth model, and Completion assesses the extent to which points in the ground truth model are covered by the reconstructed model.}


\section{More Analysis on Depth Correction}
\begin{figure*}[p]
    \centering
    \includegraphics[width=1.0\linewidth]{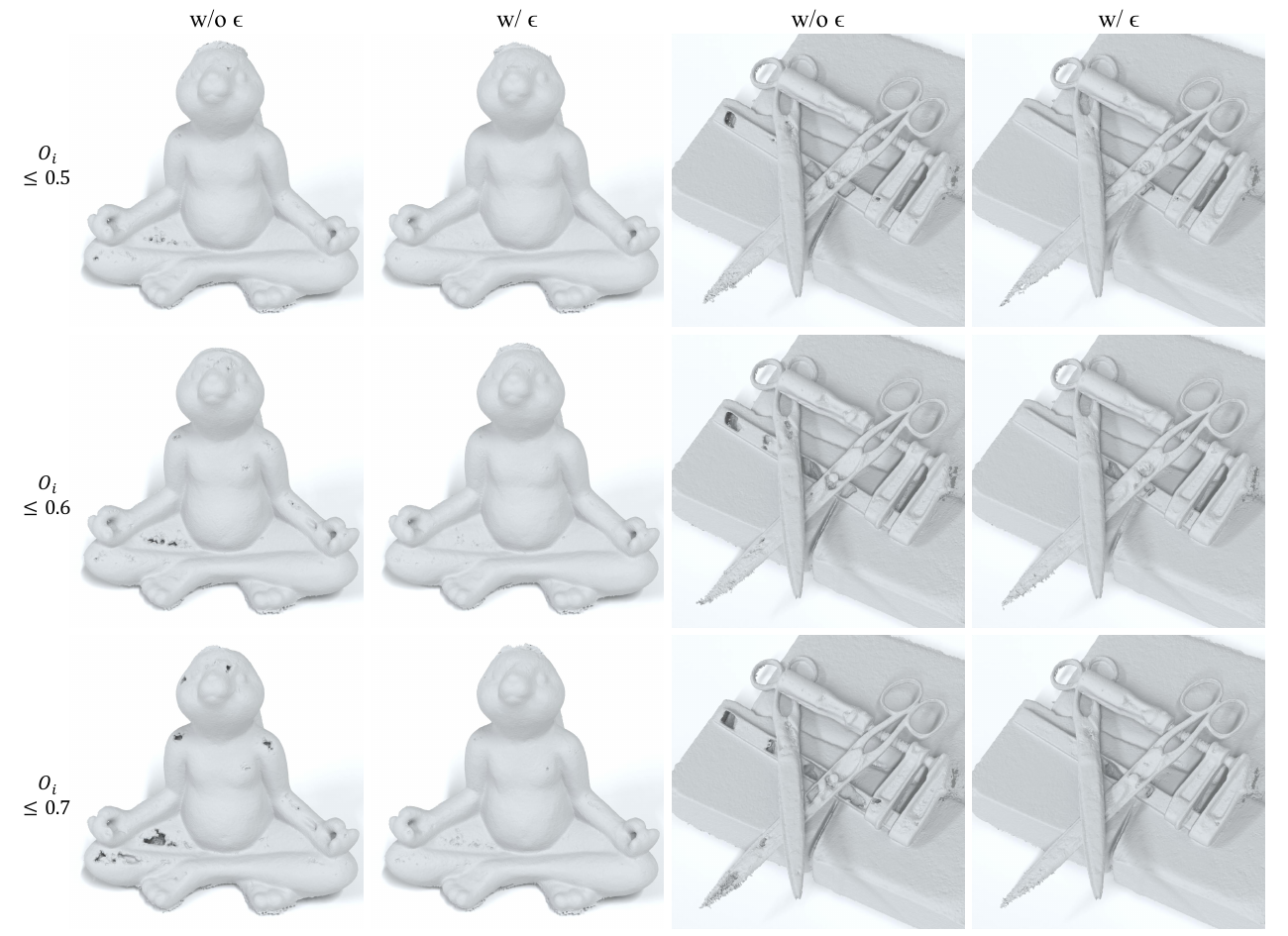}
    \caption{Ablation results of cumulative opacity on DTU dataset. It can be observed that when selecting different thresholds for cumulative opacity, adding $\epsilon$ (which we set to 0.1) improves the reconstruction quality consistently, particularly in glossy and specular areas, \yang{which strongly supports the design of our depth correction that takes into account the number of intersected Gaussians along a ray.}
    }
    \label{fig:eps_ablation}   
\end{figure*}

\yang{To validate the effect of our new depth criterion independently, we conducted two ablation studies.
First, we modify 2DGS into two variants. As the transmittance decreases from 1 to 0.5 along a ray, one variant \yyx{involves increasing} the transmittance threshold, \yyx{allowing low-opacity Gaussians located on the actual surfaces (as visualized in Fig.~\ref{fig:vis_gs}) to contribute more significantly to depth determination.} However, as reported in the 2nd row of Table~\ref{tab:depth_ablation}, such modification brings almost no improvement to the reconstruction. In contrast, the reconstruction accuracy is notably improved after replacing the depth criterion with our new design; see the 3rd row of Table~\ref{tab:depth_ablation}.

The second ablation is conducted on the $\epsilon$ in our depth correction (equation~\eqref{eq:opacity}), which is a small constant used to control the impact of the cumulative number of \yang{intersected Gaussians}. We select two challenging scenes and show the qualitative comparison (as quantitative metrics are almost the same) in Fig.~\ref{fig:eps_ablation}. We can see that with the $\epsilon$, most of the remaining small holes in the overexposure areas are fixed by our full criterion of depth correction, serving as strong evidence of our design.
}

\setlength\tabcolsep{0.5em}
\begin{table*}[h]
\caption{Ablation results of depth correction on DTU dataset~\cite{jensen2014large}. We calculated the chamfer distance(mm)$\downarrow$ for each scanned scene. It can be observed that modifying the threshold of transmittance $T_i$ (\yang{see equation~\eqref{eq:transmittance} of our paper, or } equation (19) in \yang{Appendix B of} 2DGS~\cite{huang20242d}) has a negligible impact on the reconstruction outcomes. On the other hand, the adoption of our \yang{new} depth criterion significantly improves the reconstruction quality.} 
\centering
\resizebox{\textwidth}{!}{%
\begin{tabular}{@{}lccccccccccccccccclc}
\toprule
 \multicolumn{3}{c}{} & 24 & 37 & 40 & 55 & 63 & 65 & 69 & 83 & 97 & 105 & 106 & 110 & 114 & 118 & 122 & & Mean \\ \midrule

 & 2DGS ($T_i > 0.5$) && 0.46 & 0.81 & 0.32 & 0.37 & 0.99 & 0.89 & 0.79 & 1.32 & 1.23 & 0.68 & 0.68 & 1.41 & 0.38 & 0.66 & 0.47 & & 0.76 \\
 
  & 2DGS ($T_i > 0.6$) &&  0.47 & 0.82 &  0.33 & 0.36 & 0.9 &  0.86 & 0.79 & 1.36 & 1.25 & 0.72 & 0.70 &  1.29 & 0.38 & 0.72 &  0.49 &&  0.76 \\
  
  & 2DGS + depth\_corr && 0.46 & 0.80 & 0.36 & 0.36 &  0.88 & 0.78 & 0.72 & 1.37 & 1.21 & 0.69 & 0.63 & 1.21 & 0.38 & 0.62 & 0.48 & & 0.73 \\
 \bottomrule
\end{tabular}
}
\label{tab:depth_ablation}
\end{table*}

\section{Additional comparisons and results}
\begin{figure*}[t]
    \centering
    \includegraphics[width=0.97\linewidth]{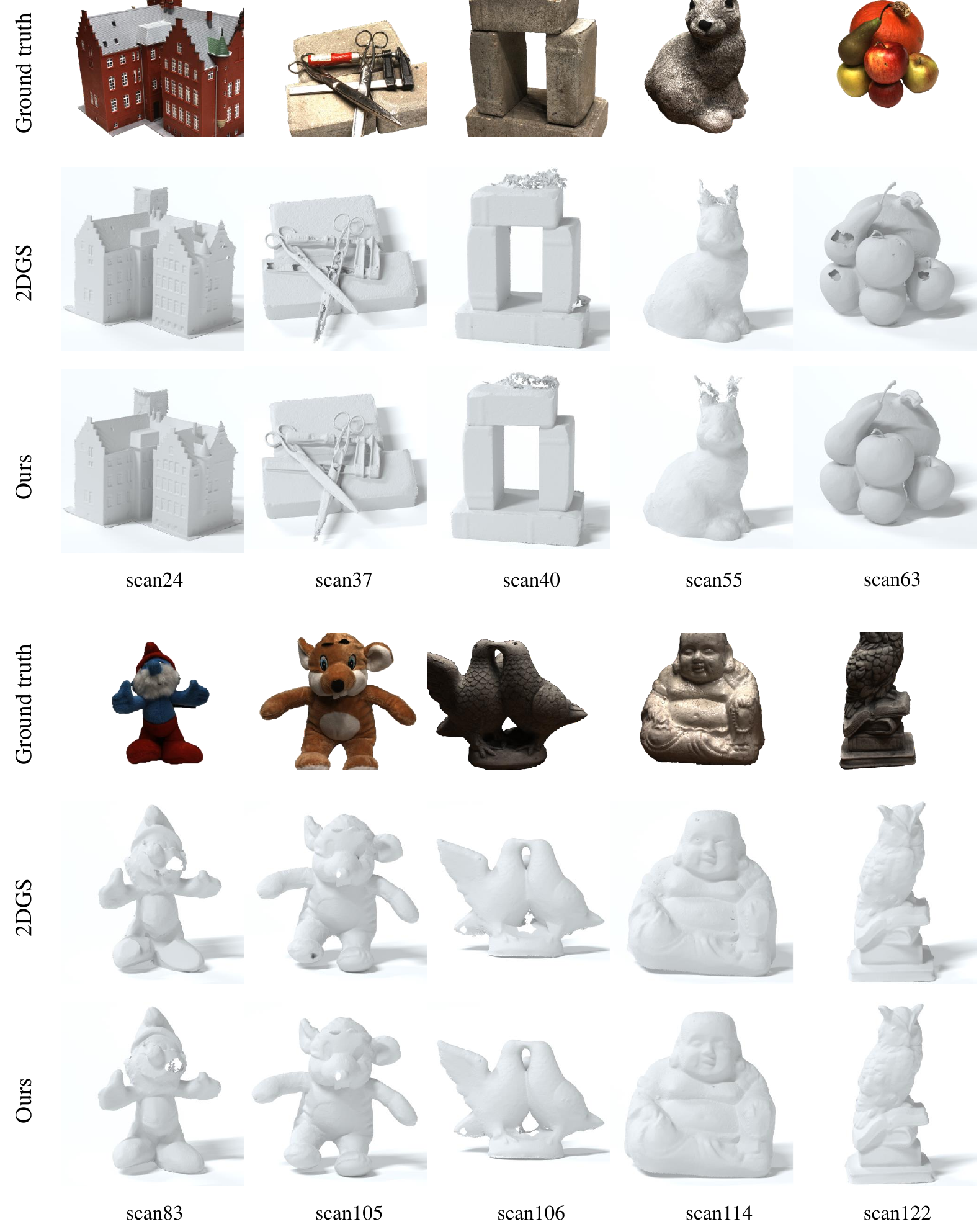}
    \caption{\yang{Additional qualitative comparisons with 2DGS on DTU dataset.} 
    }
    \label{fig:dtu_suppl}   
\end{figure*}

\begin{figure*}[t]
    \centering
    \includegraphics[width=0.95\linewidth]{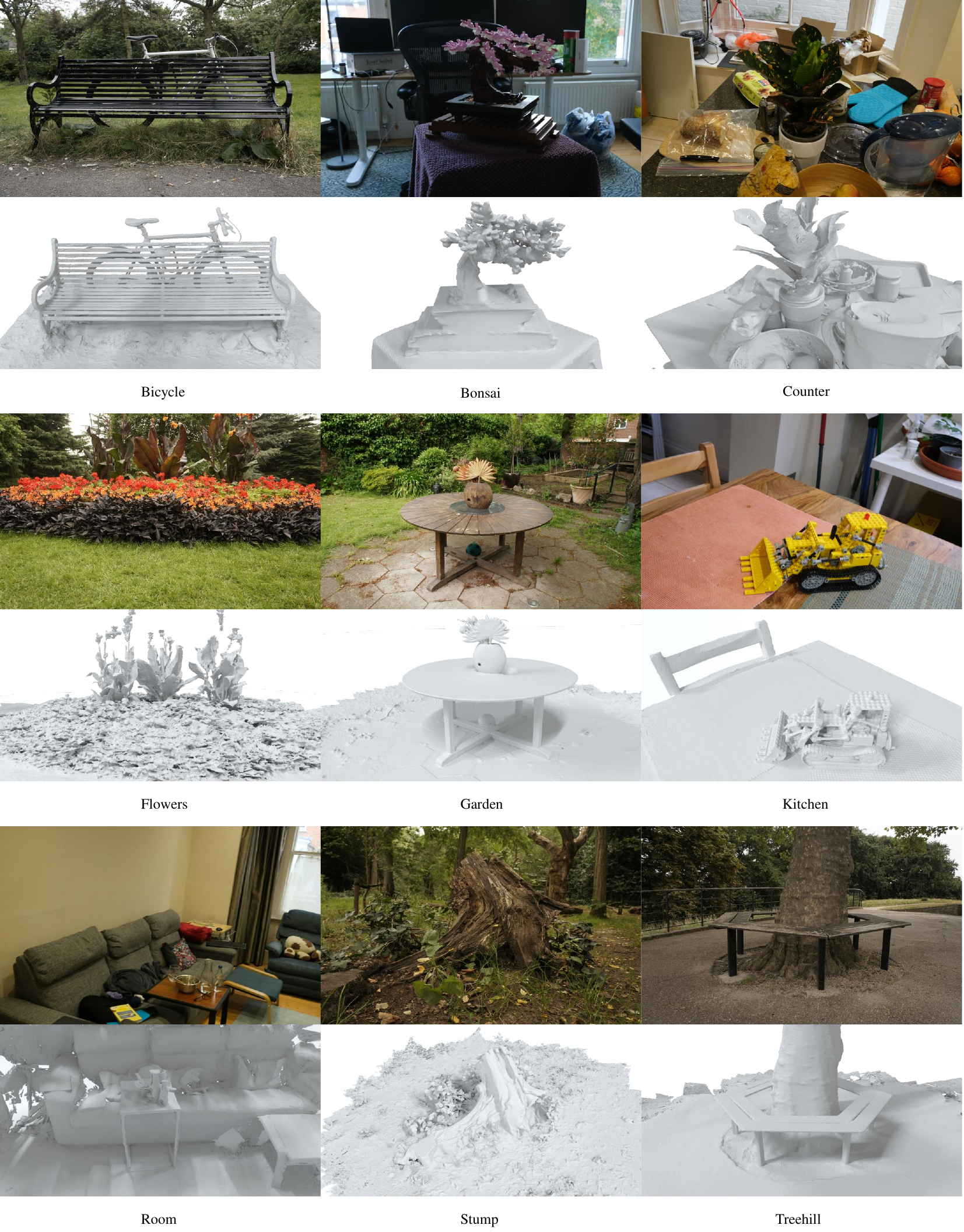}
    \caption{\yang{More geometric reconstruction results of our method on} Mip-NeRF360 dataset. 
    }
    \label{fig:mipnerf_suppl}   
\end{figure*}

\yang{
Fig.~\ref{fig:dtu_suppl} shows the additional qualitative comparison with 2DGS on the DTU dataset. Similar to Fig.~\ref{fig:dtu_results} of our main paper, holes caused by specular highlights of glossy material are effectively solved by our method. Yet, we can see holes \yyx{remaining} on the the furry surfaces of scenes ``scan83'' and ``scan105''. We attribute them to the same limitation (local dense and tiny parts) as discussed in our main paper.

Fig.~\ref{fig:mipnerf_suppl} show more qualitative results of our method on outdoor scenes of Mip-NeRF360 dataset~\cite{barron2022mip}.
}

\end{document}